\pdfoutput=1

\documentclass[11pt]{article}

\usepackage[preprint]{acl}

\usepackage{times}
\usepackage{latexsym}

\usepackage{booktabs}%

\usepackage{etaremune}

\usepackage[T1]{fontenc}

\usepackage[utf8]{inputenc}

\usepackage{microtype}

\usepackage{inconsolata}

\usepackage{graphicx}

\usepackage{enumitem}

\usepackage{tabu}
\usepackage{multirow}
\usepackage{booktabs}
\usepackage{arydshln}

\usepackage{caption} 
\usepackage{subcaption}

\newcommand{\xref}[1]{\S\ref{#1}}

\title{Beyond Turn-Based Interfaces: \\ Synchronous LLMs as Full-Duplex Dialogue Agents}

\newcommand{\squishlist}{\begin{itemize}[itemsep=1pt,parsep=2pt,topsep=3pt,partopsep=0pt,leftmargin=0em, itemindent=1em,labelwidth=1em,labelsep=0.5em]}
\newcommand{\squishend}{\end{itemize}}

\author{
Bandhav Veluri$^{1,2}$, Benjamin N Peloquin$^1$,  Bokai Yu$^1$, \\
{\bf Hongyu Gong$^1$,} {\bf Shyamnath Gollakota$^2$} \\
{ $^1$Meta AI, $^2$University of Washington} \\
\texttt{\{bandhav,gshyam\}@cs.washington.edu\, hygong@meta.com}}

\begin{document}

\maketitle

\begin{abstract}

Despite broad interest in modeling spoken dialogue agents, most approaches are inherently ``half-duplex’’ -- restricted to turn-based interaction with responses requiring explicit prompting by the user or implicit tracking of interruption or silence events. Human dialogue, by contrast, is ``full-duplex’’ allowing for rich synchronicity in the form of quick and dynamic turn-taking, overlapping speech, and backchanneling. Technically, the challenge of achieving full-duplex  dialogue with LLMs lies in modeling synchrony as pre-trained LLMs do not have a sense of ``time’’.  To bridge this gap, we propose Synchronous LLMs for full-duplex spoken dialogue modeling. We design a novel mechanism to integrate time information into Llama3-8b so that they run synchronously with the real-world clock. We also introduce a training recipe that uses 212k hours of synthetic spoken dialogue data generated from text dialogue data to create a model that generates meaningful and natural spoken dialogue, with just 2k hours of real-world spoken dialogue data. Synchronous LLMs outperform state-of-the-art   in dialogue meaningfulness while maintaining naturalness. Finally, we demonstrate the model's ability to participate in full-duplex dialogue by simulating interaction between two agents trained on different datasets, while considering Internet-scale latencies of up to 240ms. Webpage: \textcolor{blue}{\url{https://syncllm.cs.washington.edu/}}.

\end{abstract}

\section{Introduction}

Existing spoken dialogue models are predominantly turn-based interfaces that are half-duplex in nature~\cite{lakhotia2021generative, speechgpt, hassid2024textually, borsos2023audiolm}. To achieve a change of turn, these systems rely on either explicit user inputs or pauses at the end of a user’s utterance~\cite{speechgpt}. Human spoken dialogue, by contrast, does not rely on silence as its primary turn-taking cue~\cite{10.3389/fpsyg.2015.00731, dgslm}. Research indicates that in human conversations intra-turn pauses (pauses within a speaker's turn) are usually longer than the intervals between turns across speakers~\cite{turntake3,Brady1968ASA,Bosch2005OnTA}. English speakers often begin their turns without waiting for pauses, using grammatical, prosodic, and pragmatic cues to seamlessly initiate their next turn while minimizing overlaps and gaps~\cite{Stivers2009UniversalsAC}.

Human spoken dialogue is inherently full-duplex, allowing for seamless, bi-directional communication where both parties can simultaneously speak and listen. This mode of interaction enables immediate feedback, interruptions for clarification, and real-time adjustments in information flow~\cite{doi:10.1126/sciadv.adf3197, frontiers}.
Unlike half-duplex systems that process text or speech based on full utterances in each turn, human dialogue frequently contains verbal backchannels -- short, overlapping phrases such as "yeah" or "uh-huh" -- signals from the listener to the speaker that they understand and that the speaker may continue. Such synchronous dynamics allow the interaction to flow smoothly and create a rhythm absent in written text~\cite{turntake3}.
While humans learn turn-taking cues from infancy to minimize speech overlaps and silence duration~\cite{Nguyen2021ASR}, overlapping speech as well as  long silences are common in human spoken dialogue as they enrich conversations providing additional pragmatic cues. For example, overlapping speech and frequent backchanneling often signifies engaged listening. Similarly the length of silences can vary across cultures and is influenced by the promptness of responses~\cite{Stivers2009UniversalsAC,dgslm}. In both cases, these dynamics make conversation sound more ``human.''



\begin{figure*}[!ht]
\centering
\includegraphics[width=0.79\linewidth]{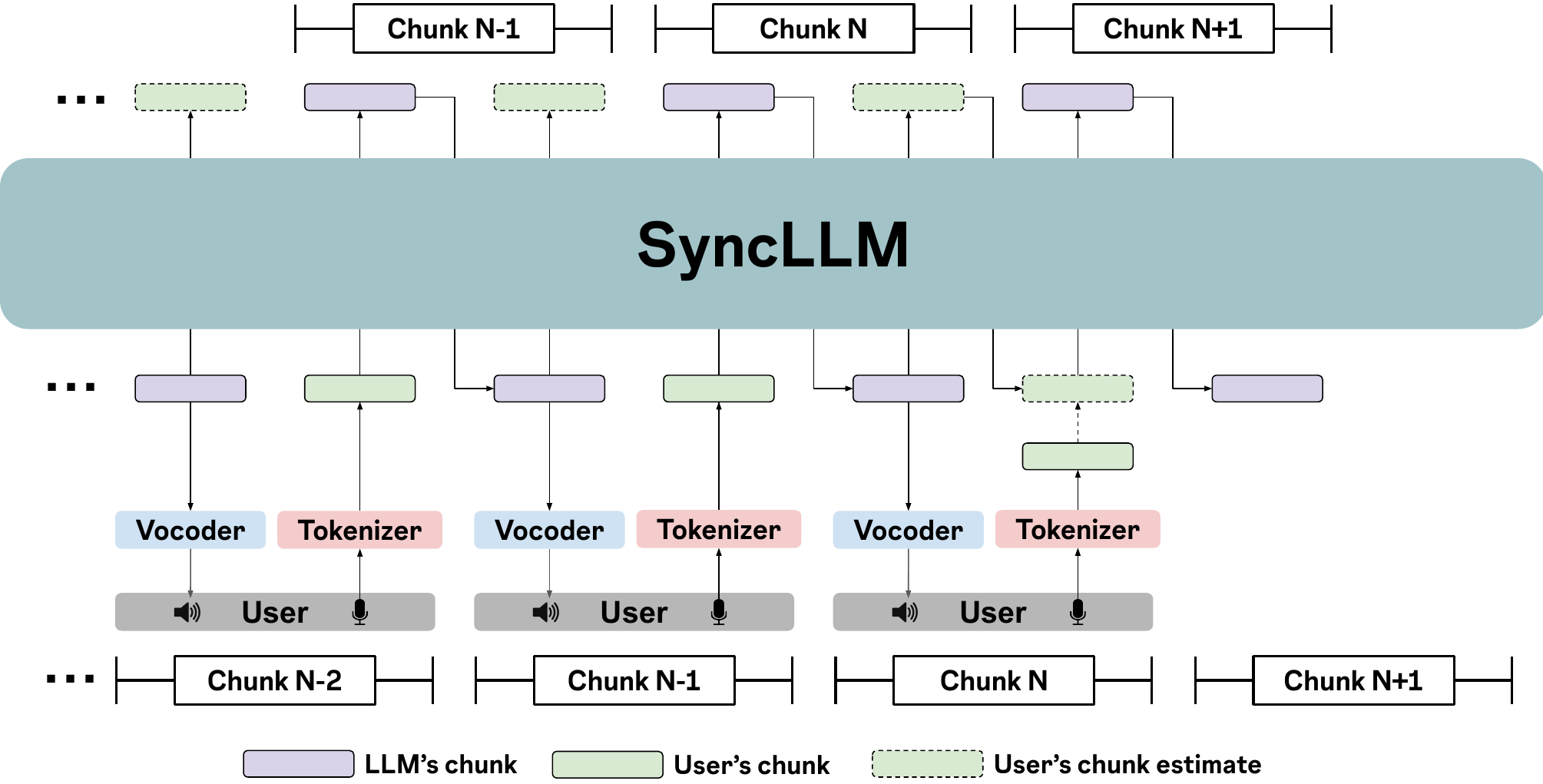}
\caption{\textbf{SyncLLM as a full-duplex dialogue agent.} At current time step (chunk N in the figure), SyncLLM's context contains interleaved chunks of the  LLM's speech until the current chunk, and the user's speech corresponding to all but the current chunk. To be in synchrony with the user, the LLM must generate its next chunk (chunk N+1) before the end of the current chunk. As a result, SyncLLM first generates an \emph{estimated user's chunk}, which is in-turn appended to the context and used to predict its next chunk.}
\label{fig:interaction}
\vskip -0.15in
\end{figure*}

Developing a full-duplex spoken dialog agent is challenging for four reasons: 1) Understanding and generating turn-taking cues in spoken dialogue requires the model to have a common reference clock with the real-world. However, current LLMs do not have such a sense of ``time''. 2) Compared to text-based chat datasets, spoken dialogue data is limited. A combination of all significant spoken dialogue datasets \cite{Cieri2004TheFC, 225858, doi:10.1126/sciadv.adf3197} would still result in only $\sim$3k hours of spoken dialogue data. 3) Full-duplex dialogue entails model to be always listening and should always be ready to speak, because back-channels or overlaps could occur at arbitrary points in time. This requires the model to be streaming for the duration of the dialogue. 4) Since the spoken dialogue agent might run on cloud infrastructure, it must address the fundamental latency inherent in Internet transmissions. Thus, the model may not have immediate access to the current tokens or speech generated by the user and must operate with delayed input (Fig.~\ref{fig:interaction}).

In this paper, we make multiple  contributions to  develop a full-duplex dialogue agent:
\squishlist
    \item We introduce Synchronous LLMs, in short SyncLLM, for full-duplex spoken dialogue. SyncLLM achieves synchrony modeling by 
    integrating time information into LLMs so that they can run synchronously with the real-world clock. We generate a periodic synchronization token to provide a common time frame for both sides of the dialogue.  This however requires us to address duplicate  tokens, caused by silence within and across utterances. Duplicate tokens can  adversely affect the semantic capability of spoken dialogue model~\cite{dgslm}. Instead, we train our  model to predict deduplicated token sequences, with  timing information maintained by our periodic synchronization tokens.
    \item Human voice interactions rely on the ability to model the other person's response on the short-term. We can take turns with gaps as small as 200ms, while language generation latency is around 600ms~\cite{frontiers}. This implies  we anticipate the next few words of what the other person would  say and respond appropriately. We use this insight  to predict speech units for  both speakers, into the future, in chunk sizes of 160-240~ms. This  ensures resiliency  to Internet latencies  of up to 240~ms.
    \item We propose a three-stage training recipe that leverages synthetic spoken dialogue generated from text dialogue data to mitigate the limited availability of real-world spoken dialogue data. Specifically, we use 212k hours of synthetic spoken dialogue data and just 2k hours of real-world spoken dialogue data to develop a model that  generates meaningful spoken dialogue with naturalistic turn-taking, overlaps, and backchannels. 
    \item With an experimental setup based on Llama3-8b \cite{llama3} and extensive user-study (n=32), we show that our method achieves +2.2-point Mean Opinion Score (MOS) improvement in dialogue content \textit{Meaningfulness} over state-of-the-art full-duplex voice model dGSLM \cite{dgslm}, while maintaining turn-taking \textit{Naturalness}. Further, our results show that our model fine-tuned on the Fisher training set~\cite{Cieri2004TheFC} can generalize to the out-of-distribution  Candor testset~\cite{doi:10.1126/sciadv.adf3197}, while preserving both dialog content meaningfulness and naturalness.
    \item Finally, by simulating full-duplex dialogue between two finetuned Llama3-8b models, we show how this approach can enable latency-tolerant and streaming full-duplex voice interfaces. Further, SyncLLM can perform a coherent conversation  even when the user’s side of the conversation is generated by a model trained with a different dataset.
\squishend

\section{Related work}


\noindent\textbf{Multimodal language models}. The success of text language models like GPT-4 \cite{gpt4}, \textsc{Llama} \cite{llama2}, and Mistral \cite{mistral} has inspired explorations into multimodal models. Here, we focus our discussion on speech and text modalities. Initialization from a pretrained text LLM has been shown to benefit multimodal training \cite{textually_pretrain}. Recent works have proposed extending the vocabulary of text LLMs with discrete speech tokens to enable the model to handle speech inputs and outputs \cite{rubenstein2023audiopalm}. Models are trained with cross-modal knowledge from aligned speech-text data, including tasks like automatic speech recognition (ASR), text-to-speech synthesis (TTS), speech-to-text (S2T), and speech-to-speech translation (S2ST). Multitask learning with these tasks has been adopted by  VioLA \cite{viola}, AudioPaLM \cite{rubenstein2023audiopalm}, VoxtLM \cite{voxtlm}, and SUTLM \cite{sutlm}. SpiRit-LM \cite{nguyen2024spiritlm} interleaves speech and text tokens and trains the model with next token prediction, demonstrating both speech understanding and generation.



\noindent\textbf{Spoken dialogue models}. Prior work on spoken dialogue research covers various topics such as dialogue state tracking \cite{monet}, turn-taking prediction \cite{turn_taking, Lin_2022}, and response generation \cite{zhang2019dialogpt}. Recent works leverage LLMs in dialogue systems \cite{dialogue_gen}.
Initialized from \textsc{Llama}, SpeechGPT \cite{speechgpt} is finetuned sequentially on speech-only data and multimodal instruction sets to perform spoken question answering (QA) tasks. USDM \cite{USDM} continues pretraining Mistral with interleaved speech-text data to capture multimodal semantics. For dialogue finetuning, it constructs templates using both speech and transcripts of user input as instruction data. Unlike models that use speech tokens, Spectron \cite{spectron} directly manipulates spectrograms for tasks such as spoken QA and speech continuation. However, these prior works are limited to the turn-taking setting, where the dialogue model is explicitly prompted to speak in its own turn. Human spoken dialogue is more complex, involving implicit turn-taking cues and overlapping speech, such as interruptions and backchanneling \cite{schegloff2000overlapping}. 

The closest work to ours is dGSLM \cite{lakhotia2021generative}, which models simultaneous dialogue using a dual-tower Transformer that attends to two channels. 
It demonstrates superior performance than cascaded architecture which consists of automatic speech recognition (ASR), text LLM and text-to-speech (TTS). 
One weakness of dGSLM is its reliance on speech-only training, which does not fully utilize textual knowledge. In contrast, our work leverages the generative intelligence of language models, equipping them with multimodal and synchronous capabilities. Moreover, in its empirical study, dGSLM does not consider delays in real-life scenarios and assumes that the hidden states of one interlocutor are immediately accessible to the other. In contrast, we explicitly discuss how our model handles delayed responses in spoken dialogue.


\section{SyncLLM}


SyncLLM is an auto-regressive transformer decoder architecture, that natively models discrete speech units in a wall-clock synchronous fashion. SyncLLM is trained to predict interleaving chunks of speech units corresponding to both sides of the dialogue as shown in  Fig.~\ref{fig:interaction}. In each time step, the model predicts speech units corresponding to a fixed duration, referred to as the model's \emph{chunk size}, for its side of the dialogue followed by speech units corresponding to user's side of the dialogue. With this approach, the model is capable of generating two streams of speech synchronized with a real-world clock. This allows our method to model all conversational cues such as backchannels, overlaps, interruptions etc. Furthermore, since we use the same architecture as  existing LLMs, our approach can  leverage large scale pre-training of LLMs.

The model trained to predict interleaved chunks of token sequences can be used for full-duplex voice interaction if we could replace one of the two token streams, with that corresponding to the real-world user. In Fig.~\ref{fig:interaction}, purple boxes correspond to token sequences of the LLM's side of the conversation in each time chunk and the green boxes correspond to the user's side of the dialogue. We  achieve full-duplex LLM-user voice interaction by discarding the LLM's predictions of user's response and replace it with the user's speech.

\begin{figure*}[!ht]
\centering
\includegraphics[width=0.95\linewidth]{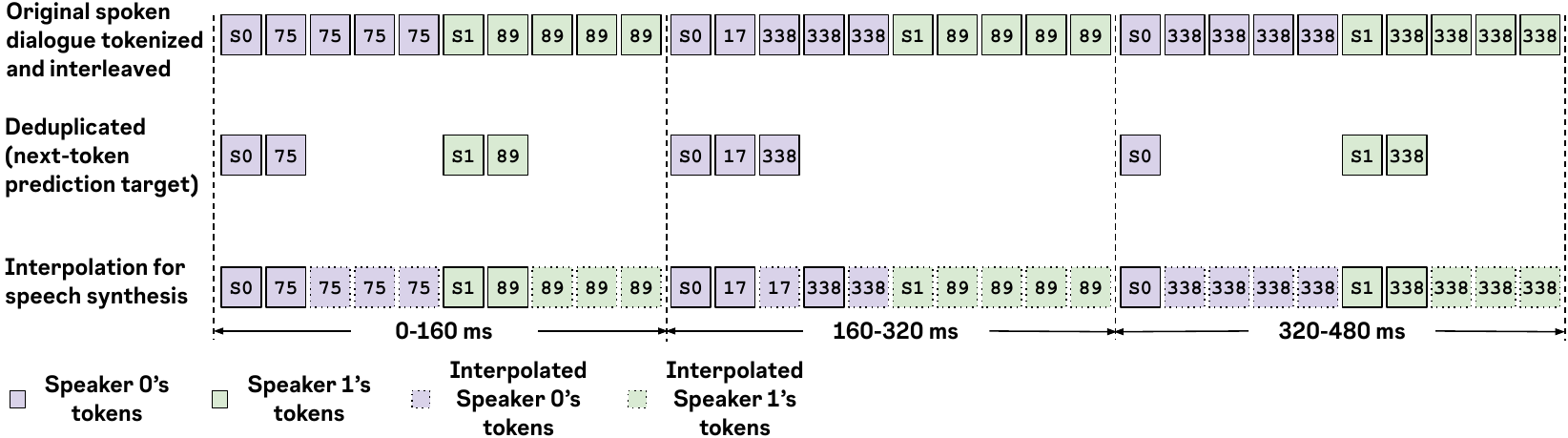}
\caption{SyncLLM's token sequence format visualized with a chunk size of 160 ms. (Top row) We represent spoken dialogue as interleaved chunks of HuBERT tokens, where the chunk size determines the frequency of the synchronization token [S0]. (Middle row) We train SyncLLM to generate interleaved chunks of deduplicated HuBERT tokens along with periodic synchronization tokens. (Bottom row) We interpolate deduplicated tokens in each chunk to obtain spoken dialogue sequence in the original format.}
\label{fig:token_sequence}
\vskip -0.15in
\end{figure*}

\subsection{Latency tolerant interaction}
\label{sec:latency_tolerant}
In Fig.~\ref{fig:interaction}, consider the Nth time chunk to be current time step. We could interleave the LLM's output speech chunks until the Nth chunk, with the user's input chunks corresponding to only N-1 chunks. The reasoning here is that  the user's input for the Nth chunk is not available until the end of Nth time step. To handle this intrinsic latency, similar to the way humans anticipate the next few words of what the other person taking part in the dialogue would say~\cite{frontiers}, the LLM's output for the next chunk (N+1) is  computed by first estimating the user's response for the Nth time chunk (depicted in the figure with green boxes with dotted border). We then append this estimated chunk to the LLM's context to generate the LLM's next chunk (N+1). For generating subsequent chunks (N+2, N+3, ...), we discard the estimated user's chunk for Nth time step and replace that with the user's real-world input, thus grounding the subsequent interaction with actual input from the user.

\subsection{Token sequence format}

Following prior works in spoken language modeling \cite{dgslm, nguyen2024spiritlm}, we use HuBERT \cite{hsu2021hubert} to represent speech. We use the tokenization parameters from \cite{nguyen2024spiritlm}, with a token sampling rate of 25 Hz -- resulting in one token for every 40 ms of audio -- and a vocabulary size of 501. To model dialog between two speakers 0 \& 1, we define two special tokens \verb|[S0]| and \verb|[S1]|, referred to as speaker tags, specifying the start of each speaker's token sequence, respectively. We represent dialogue as two parallel speech streams, one for each speaker, interleaved, as shown in the top row of Fig. \ref{fig:token_sequence}. For each stream, we embed a periodic speaker tag, with the time period equal to chunk size of the model. 

\begin{figure}[t]
\centering
\vskip -0.05in
\includegraphics[width=0.85\linewidth]{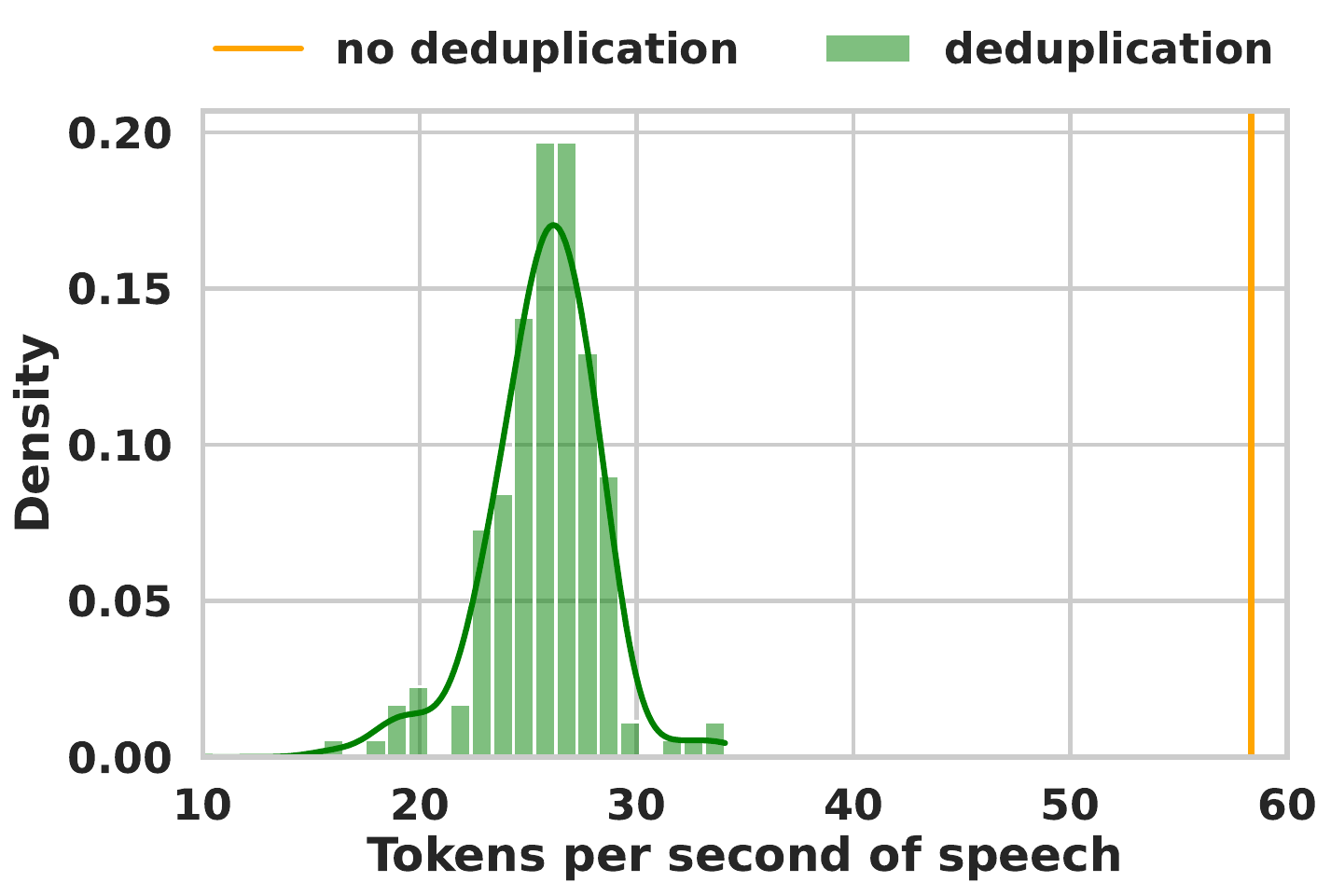}
\vskip -0.1in
\caption{Tokens required for representing a second of speech with/without deduplication. Histogram computed over 15 hr of  dialog data in the Fisher dataset \cite{Cieri2004TheFC}.}
\vskip -0.2in
\label{fig:deduplication}
\end{figure}

\textbf{Deduplication.} The fixed time period of HuBERT tokens is useful for modeling time in the full-duplex dialogue. However, raw HuBERT sequences consist of significant repeated tokens, mainly caused by silence within and across utterances. The number of repetitions of each unique token denote the duration of the acoustic unit represented by the token. The semantic content, however, can be modeled by only considering unique tokens while deduplicating the token sequence \cite{kharitonov2022textfree, dgslm}. Duplicate token sequences can adversely affect the semantic capability of the final spoken dialogue model \cite{dgslm}, because as shown in Fig. \ref{fig:deduplication}, they contain $\sim50\%$ lower semantic content per token compared to deduplicated  sequences.

So, instead, SyncLLM is trained to predict deduplicated HuBERT sequences, with coarse timing information maintained by periodically interleaved special tokens, \verb|[S0]| and \verb|[S1]|, as in the second row of Fig.~\ref{fig:token_sequence}. In the first chunk of the example  in Fig.~\ref{fig:token_sequence}, the two speaker streams contained 4 repetitions of \verb|[75]| and \verb|[89]|, respectively. After deduplication, the interleaved token sequence corresponding to the first chunk would be \verb|[S0][75][S1][89]|. In the second chunk, speaker 0 has 2 new tokens (\verb|[17]| \& \verb|[338]|), but speaker 1 tokens are just a repetition of the last token in the previous chunk, \verb|[89]|. So, the second chunk's token sequence would just be \verb|[S0][17][338]|. Note that when a chunk contains no novel tokens corresponding to speaker 1, we exclude speaker 1's special token \verb|[S1]| as well. However, this is not the case for speaker 0, as we need one of the speaker's special token to be present in all chunks to unambiguously distinguish chunks. This is shown in the third chunk of Fig.~\ref{fig:token_sequence}. 

\textbf{Interpolation.} While deduplicated token sequences are beneficial for auto-regressive modeling, to generate token sequences suitable for speech synthesis, we need periodic HuBERT tokens in the original format. Since the speaker tag \verb|[S0]| maintains the timing information, we know the number of tokens removed after deduplication within each chunk. We use this  to interpolate the deduplicated token to match the expected number of token in each chunk. For example, in the first chunk of Fig.~\ref{fig:token_sequence}, speaker 0's stream only has one token after deduplication. But since chunk size in that case is 160ms, each chunk would contain 160/40 = 4 tokens. So as shown in the third row of Fig.~\ref{fig:token_sequence}, we repeat the deduplicated token thrice to reconstruct the chunk. If a chunk has multiple deduplicated tokens, like the second in Fig. \ref{fig:token_sequence}, we repeat each token by an equal amount. We note this approach could result in an error because the original chunk may not follow this heuristic. We observed that the effect of this  is imperceptible even with a chunk size of 240 ms,  likely because the error in the predicted duration of each token is upper bounded by the chunk size. Further, in chunks with more novel tokens, the error would be even smaller.

\section{Training}
\label{sec:method}

We use Llama3-8b \cite{llama3} as our base model and employ a three stage training procedure that uses synthetic spoken dialogue data predominantly and relatively small amount of real-world spoken dialogue data to develop a full-duplex voice agent. 


\begin{table}[!t]
\centering
\caption{
Data used for training in different stages. We convert text based data to speech using TTS.}
\label{tab:your_label}
\begin{tabular}{lccc}
\hline
\textbf{} & \textbf{Stage} & \textbf{Source} & \textbf{Speech} \\
&& \textbf{modality} & \textbf{(hrs)}\\
\hline
Supervised & 1 & Text & 193k \\
finetuning (SFT) \\
Dialogue & 2 & Text & 20k \\
Spoken dialogue & 3 & Speech & 1927 \\
\hline
\end{tabular}
\vskip -0.1in
\end{table}

\noindent\textbf{Stage 1: Turn-based spoken dialogue model with synthetic speech data.}
Given the limited spoken dialogue data, we generate synthetic speech data from large-scale text dialogue datasets.
We use supervised finetuning (SFT) datasets, as our source text-dialogue datasets. We used Bark TTS \cite{barktts} model to generate spoken versions of text-dialogue datasets, with its 10 speaker presets. 

Since Llama3-8b is a text-only LLM, in the first stage, we aim to achieve text-speech alignment in the context of dialogues. Given a spoken question, we train the model to generate a spoken response. We  expand the vocabulary of Llama3 to include 501 HuBERT tokens, in addition to the speaker tags, \verb|[S0]| and \verb|[S1]|. A turn-based dialog could be defined as made of turns, which in turn are made of sentences. We finetuned Llama3 with dialog sequences in the following format:

\begin{verbatim}
[S1]<sent0>[S0]<sent0><sent1>[S1]..
\end{verbatim}

\begin{figure}[t]
\centering
\includegraphics[width=0.7\linewidth]{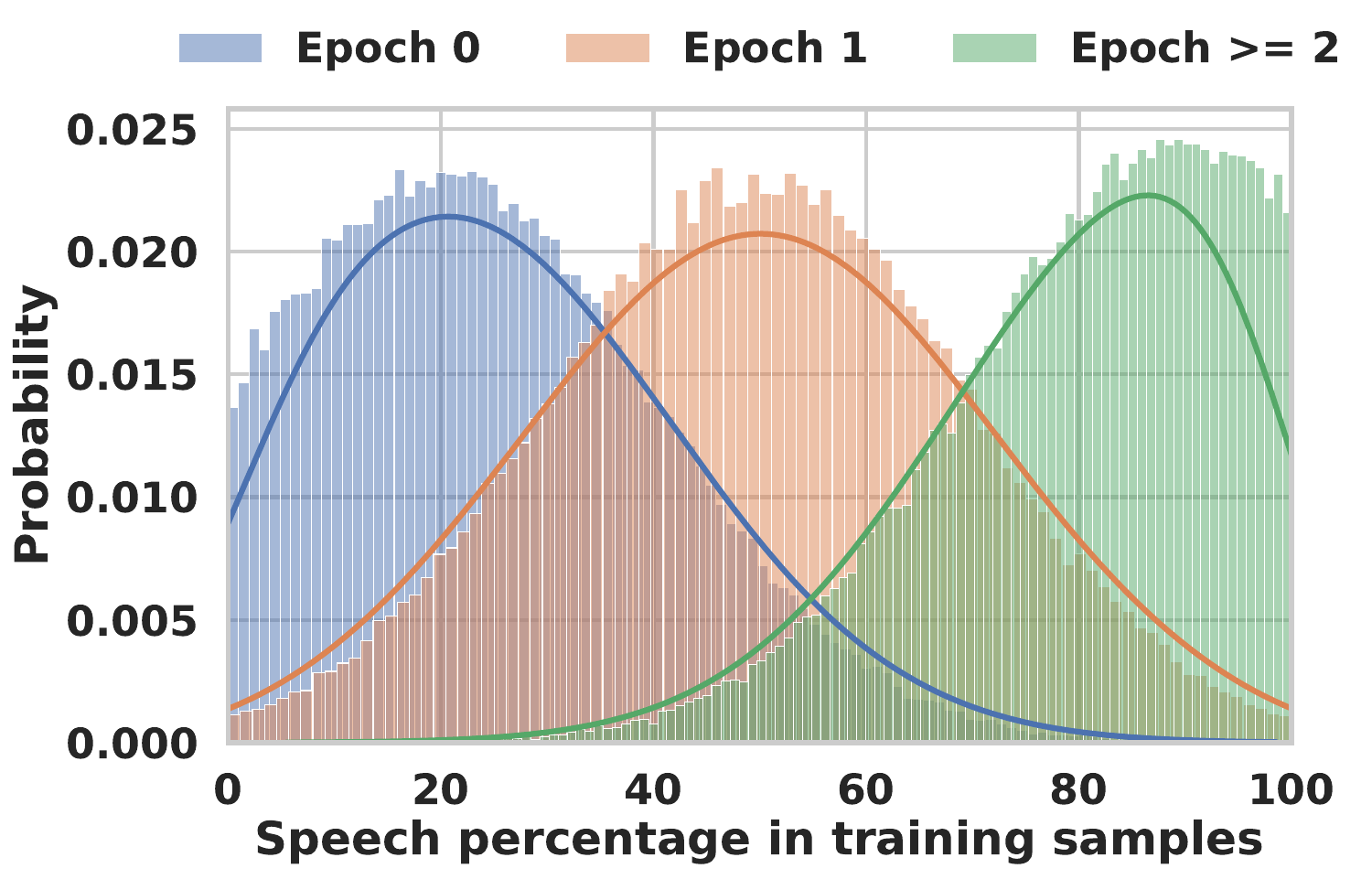}
\vskip -0.1in
\caption{We sample speech percentages from truncated normal distribution, so we obtain samples with all possible combinations of text-speech interleaving throughout the training process, with a bias for higher speech percentages as the training progresses. This  resulted in stable training when starting out with a text-only LLM.}
\label{fig:truncnorm}
\vskip -0.15in
\end{figure}

Each sentence is randomly chosen to either be text or deduplicated speech token sequences during training. For each training sample, we sample the percentage of speech sentences in the training sequence from the truncated normal distribution   (Fig.~\ref{fig:truncnorm}). Training only with fully speech sequences or step-wise increment of speech percentage resulted in unstable training. Sentence level text-speech interleaving  not only trains the model to be capable of performing dialog, but also achieves text/speech alignment in the context of dialog. 

\noindent\textbf{Stage 2: Full-duplex dialogue assuming no overlaps.} Turn-based spoken dialogue is special case of full-duplex dialogue with no overlaps. Based on this observation, we could treat synthetic spoken dialogue data as full-duplex spoken dialogue data where during one speaker's turn, other speaker is completely silent. In this stage, we create synthetic spoken dialogue data from text-dialogue data similarly to the previous stage with one main difference: From each turn in the dialogue, we generate a speech utterance corresponding to one speaker and silence of equal duration corresponding to the other speaker. We then tokenize the parallel speech dialog data in the format shown in the second row of Fig.~\ref{fig:token_sequence}. This way, we can further leverage text-dialogue data for help our model learn the token sequence format  in Fig.~\ref{fig:token_sequence}. This stage of finetuning models timing within an utterance. The model cannot learn turn-taking cues such as back-channeling or overlaps between two speakers yet.

For the the previous stage, most samples in SFT datasetswould contain one speaker (user of the LLM) taking a short turn and the other speaker (the LLM) giving a long response. Spoken dialogues however contain more frequent turn-taking taking with short utterances. Therefore for this stage, we use text-dialogue datasets comprising of shorter turns, equivalent to around $20$k hrs of synthetic spoken dialogue.

\begin{table*}[!t]
\centering
\caption{Comparison of Pearson correlation of turn-taking event durations between generations and ground-truth continuations, given same set of prompts. SyncLLM's chunk sizes are shown in parenthesis.}
\vskip -0.1in
\label{tab:gt_corr}
\begin{tabular}{l*{4}{c}c*{4}{c}}
\hline
\textbf{Model} & \multicolumn{4}{c}{\textbf{Fisher (in-distribution)}} & \phantom{|} & \multicolumn{4}{c}{\textbf{Candor (out-of-distribution)}} \\
\cline{2-5} \cline{7-10}
& \textbf{ipu} & \textbf{pause} & \textbf{fto} & \textbf{Average} && \textbf{ipu} & \textbf{pause} & \textbf{fto} & \textbf{Average} \\
\hline
dGSLM & 0.48 & 0.41 & 0.10 & 0.33 && 0.30 & 0.02 & 0.09 & 0.14 \\
SyncLLM-F (160 ms) & 0.60 & 0.50 & 0.20 & 0.43 && 0.45 & 0.09 & 0.14 & 0.23 \\
SyncLLM-F (200 ms) & 0.60 & 0.49 & 0.19 & 0.43 && 0.44 & 0.28 & 0.14 & 0.29 \\
SyncLLM-F (240 ms) & 0.58 & 0.40 & 0.25 & 0.41 && 0.45 & 0.27 & 0.21 & 0.31 \\
\hline
Prompt & 0.72 & 0.53 & 0.31 & 0.52 && 0.54 & 0.30 & 0.12 & 0.32 \\
Resynth-GT & 0.92 & 0.92 & 0.53 & 0.79 && 0.90 & 0.86 & 0.37 & 0.71 \\
\hline
\end{tabular}
\vskip -0.15in
\end{table*}

\noindent\textbf{Stage 3: Modeling with real-world spoken dialogue data.} Finally, we finetune the model to learn turn-taking cues from real-world spoken dialogue data. We use the Fisher \cite{Cieri2004TheFC} dataset with  2000 hours of spoken dialogues, where each speaker's speech in a dialogue is separated into independent audio channels. {We split the dataset into train, val and test split with 98:1:1 ratio, respectively.} Each audio channel in the dialogue is separately tokenized and interleaved in the full-duplex dialogue format used in the previous stage. In this stage in addition to learning timing within utterances, the model learns effective turn-taking, conversational cues like accurate distribution of pauses between turn and backchanneling.


\section{Experiments}
\label{sec:experiments}


\begin{figure}[t!]
\centering
\includegraphics[width=0.85\linewidth]{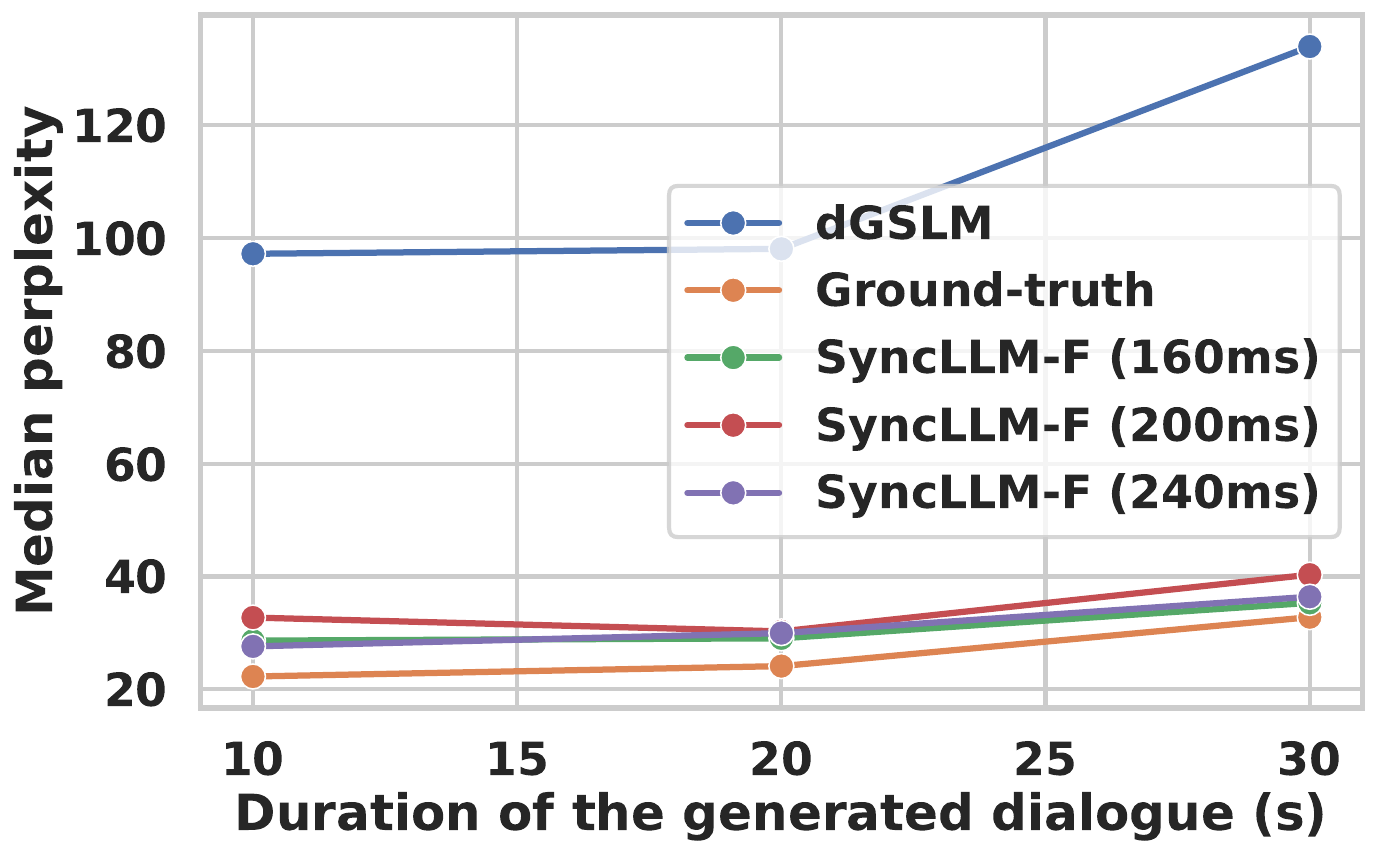}
\vskip -0.05in
\caption{Perplexity of transcriptions of spoken dialogues generated by different models. Perplexity is measured with respect to a text dialogue model's predictions.}
\label{fig:cont_comp}
\vskip -0.15in
\end{figure}

{We evaluate SyncLLM in both continuation and interaction settings. In the continuation setting, given a spoken dialogue prompt, the model generates both sides of the dialogue. For interaction setting, we simulate interaction between two instances of SyncLLM as described in \xref{sec:latency_tolerant}.  We denote SyncLLM trained on Fisher in continuation setting as SyncLLM-F and use dGSLM as the continuation setting baseline. Both dGSLM and SyncLLM-F use Fisher as the only real-world spoken dialogue dataset for training. We denote SyncLLM trained on Fisher interacting with an instance trained on Fisher as SyncLLM-F-F, and SyncLLM trained on Fisher interacting with an instance trained on CANDOR \cite{doi:10.1126/sciadv.adf3197} as SyncLLM-F-C. }

\subsection{Semantic evaluation}
\label{sec:semantic_eval}
We evaluate the semantics of SyncLLM in the text domain by converting spoken generations to text using ASR. We transcribe the generated spoken dialogues into turn-based text dialogues ignoring any overlapping speech. We then compute perplexity of transcribed dialogues generated with 10 second spoken dialogue prompts, with respect a text-only dialogue model. To account for outliers (samples with abnormally high perplexities), we consider median perplexity over the testset. 

\begin{figure}[t]
\centering
\includegraphics[width=0.85\linewidth]{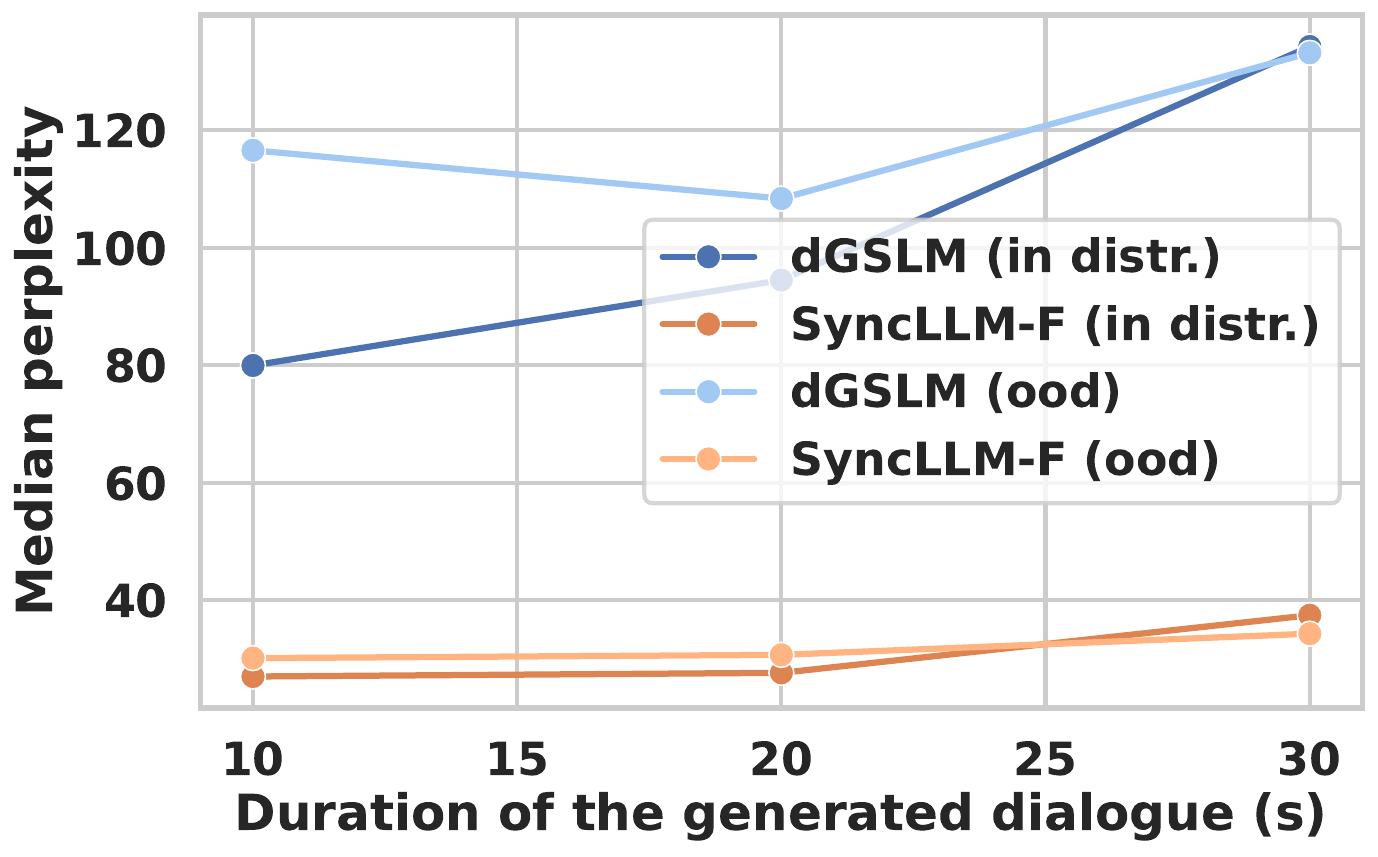}
\vskip -0.05in
\caption{In-distribution and out-of-distribution testing.}
\vskip -0.15in
\label{fig:cont_ood}
\end{figure}

Fig.~\ref{fig:cont_comp}  compares the semantic quality of spoken dialogues generated by SyncLLM  with different chunk sizes to the prior state-of-the-art full-duplex  dGSLM model~\cite{dgslm} and ground-truth continuations. We find that  dGSLM has a perplexity drop of $\sim$70 relative to the ground-truth, while SyncLLM only has a drop of $\sim$15. Fig.~\ref{fig:cont_ood} also compares median perplexities measured with prompts sampled from Fisher and Candor test splits separately, with all models trained only on Fisher training split. Here, Candor test split is an out-of-distribution testset. 

These evaluations show that our approach of using the standard auto-regressive architecture, thus leveraging vast text pre-training, results in much more semantically coherent spoken dialogue model, compared to a custom architecture proposed for speech-only training. Furthermore, our three-stage training approach leveraging large amount of synthetic spoken dialogue data generated from text dialogues, allows us to converge much faster on limited real-world dual-channel spoken dialogue data. This results in a general model that has superior out-of-distribution (ood) performance.

\subsection{Naturalness evaluation}
\label{sec:naturalness_eval}
Appropriate timing of pauses, speaker transitions and overlaps are integral part of spoken-dialogue which convey essential information required for natural spoken conversation. To evaluate these aspect of our generated spoken dialogues, we consider the turn-taking events proposed in \cite{dgslm} that evaluate overall naturalness of generated spoken dialogues: inter-pausal units (IPUs), pauses, and floor-transfer offset (FTO). {FTO is the duration of between turn-transitions, which is a combination of overlaps and gaps -- negative FTOs represent  overlaps and  positive FTOs represent gaps.}

Similar to dGSLM's setup, we use 30s prompts sampled from the test splits and generate 90s dialogues with different model configurations. We then compute pair-wise correlation of turn-taking event durations between the dialogue generations and ground-truth continuations, given the same prompt. We first compute voice activities of each side of dialogue (generated in separate audio channels) using the \verb|pyannote.audio| library \cite{Bredin2020}. We then measure   the  start and end timestamps for each turn-taking event. We measure the average duration of the turn-taking events in generated dialogues and then compute the Pearson correlation between the average durations observed in generations of different models and those in the ground-truth.

Table.~\ref{tab:gt_corr} compares this correlation with in-distribution Fisher \cite{Cieri2004TheFC} test-split and out-of-distribution Candor test-split. We observe that, generations with our models achieve better turn-taking event correlation with ground-truth continuations compared to dGSLM for both  in-distribution and out-of-distribution testsets. {In addition to this, we provide turn-taking event correlation with prompts and re-synthesized ground-truth continuations (Resynth-GT). Resynth-GT is obtained by re-synthesizing the tokenized ground-truth continuation. Resynth-GT does not perfectly correlate with ground-truth owing to variance in timing introduced by the tokenization process, and serves as a topline for our method.}

\subsection{Human Evaluation}
We conduct an evaluation study with 32 annotators  recruited via a third party vendor with the requirement that they had native-level English proficiency.


\begin{table*}[t!]
\caption{Meaningfulness (Meaning.) and Naturalness (Nat.) (scores 1-5) mean estimates and standard errors (in parentheses), aggregated overall and for Fisher and CANDOR subsets. {We use a 160ms chunk size for this study.}}
\vskip -0.05in
\centering
\begin{tabular}{lcccccc}
\toprule
 & \multicolumn{2}{c}{\textbf{Overall}} & \multicolumn{2}{c}{\textbf{Fisher}}  & \multicolumn{2}{c}{\textbf{CANDOR}}\\ 
\cmidrule(rr){2-3}\cmidrule(rr){4-5}\cmidrule(rr){6-7}
\textbf{Model} & Meaning. $\uparrow$ & Nat. $\uparrow$ & Meaning. $\uparrow$ & Nat. $\uparrow$ & Meaning. $\uparrow$ & Nat. $\uparrow$ \\
\midrule
dGSLM & 1.55 (0.06) & 3.95 (0.08) & 1.67 (0.09) & 4.21 (0.08) & 1.43 (0.08) & 3.70 (0.12) \\
SyncLLM-C & 3.40 (0.07) & 3.96 (0.06) & 3.14 (0.10) & 3.97 (0.08) & 3.66 (0.08) & 3.94 (0.08) \\
SyncLLM-F & 3.74 (0.06) & 3.90 (0.06) & 3.82 (0.08) & 3.98 (0.08) & 3.67 (0.09) & 3.82 (0.10) \\
Re-synth & 3.87 (0.06) & 4.03 (0.05) & 4.04 (0.08) & 4.14 (0.08) & 3.69 (0.07) & 3.91 (0.06) \\
\hline
GT & 4.96 (0.02) & 4.96 (0.02) & 4.96 (0.03) & 4.94 (0.04) & 4.97 (0.02) & 4.98 (0.02) \\
\bottomrule
\end{tabular}
\label{tab:human_eval_overall}
\vskip -0.15in
\end{table*}


We adapt the Mean Opinion Score (MOS) protocol (a 5-pt Likert scale) \cite{p808} to evaluate \textit{Naturalness} (N-MOS) of turn-taking and \textit{Meaningfulness} (M-MOS) of dialogue content.  For both N-MOS and M-MOS, annotators are presented with the prompt- and continuation-audio. Annotators are instructed to first read the descriptions of N-MOS and M-MOS, listen to the prompt audio, then listen to the continuation audio. Finally, they are asked to provide a rating considering the quality of the continuation audio relative to the information contained in the prompt. Each annotator assigned to a given prompt / continuation pair provides a rating for both N-MOS and M-MOS (see \xref{sec:n_mos}).

In total, $n_{annot}=32$ annotators provided ratings for $n_{items}=180$ items divided evenly between the CANDOR and Fisher datasets. Each sample received a rating from $1 - Bad, ..., 5 - Excellent$ by three unique raters. We compute item-level scores by taking the median score per item. To compute system-level scores we take the mean of item scores for a given system. We compute 95\% confidence intervals via bootstrapping, resampling at the item level for $n_b=1000$ iterations.

\noindent\textbf{Overall results.} The two left-most columns of Table.~\ref{tab:human_eval_overall} indicate that nearly all models are at parity in perceived \textit{Naturalness} (N-MOS) of turn-taking, while close to re-synthesized ground-truth values. On the perceived \textit{Meaningfulness} (M-MOS) of the dialogue content, SyncLLM-based models significantly outperform dGSLM, approaching re-synthesized ground-truth values. {Resynth-GT here accounts for the tokenization process and is the topline number for the implementation of our method using the HuBERT tokenizer.}


\noindent\textbf{In-distribution and OOD.} Table.~\ref{tab:human_eval_overall} also highlights the difference between in-distribution (Fisher) and OOD  (CANDOR) between dGSLM and Fisher-trained SyncLLM-F. While dGSLM suffers from significant degradation OOD (dropping -0.24 and -0.51 in M-MOS and N-MOS ratings), these declines are reduced in SyncLLM-F only dropping -0.15 and -0.16 moving OOD. SyncLLM trained on CANDOR dataset (SyncLLM-C) shows a  decline OOD on M-MOS (-0.52), but not N-MOS (+0.03). We note that dGSLM \cite{dgslm} uses speech representations fine-tuned on the Fisher dataset, while our method uses general-purpose speech representations for all domains of speech. This results in our method outperforming the baseline on the out-of-distribution Candor testset in naturalness, as judged by human evaluators in Table.~\ref{tab:human_eval_overall}.



\begin{figure}[!t]
\centering
\includegraphics[width=0.85\linewidth]{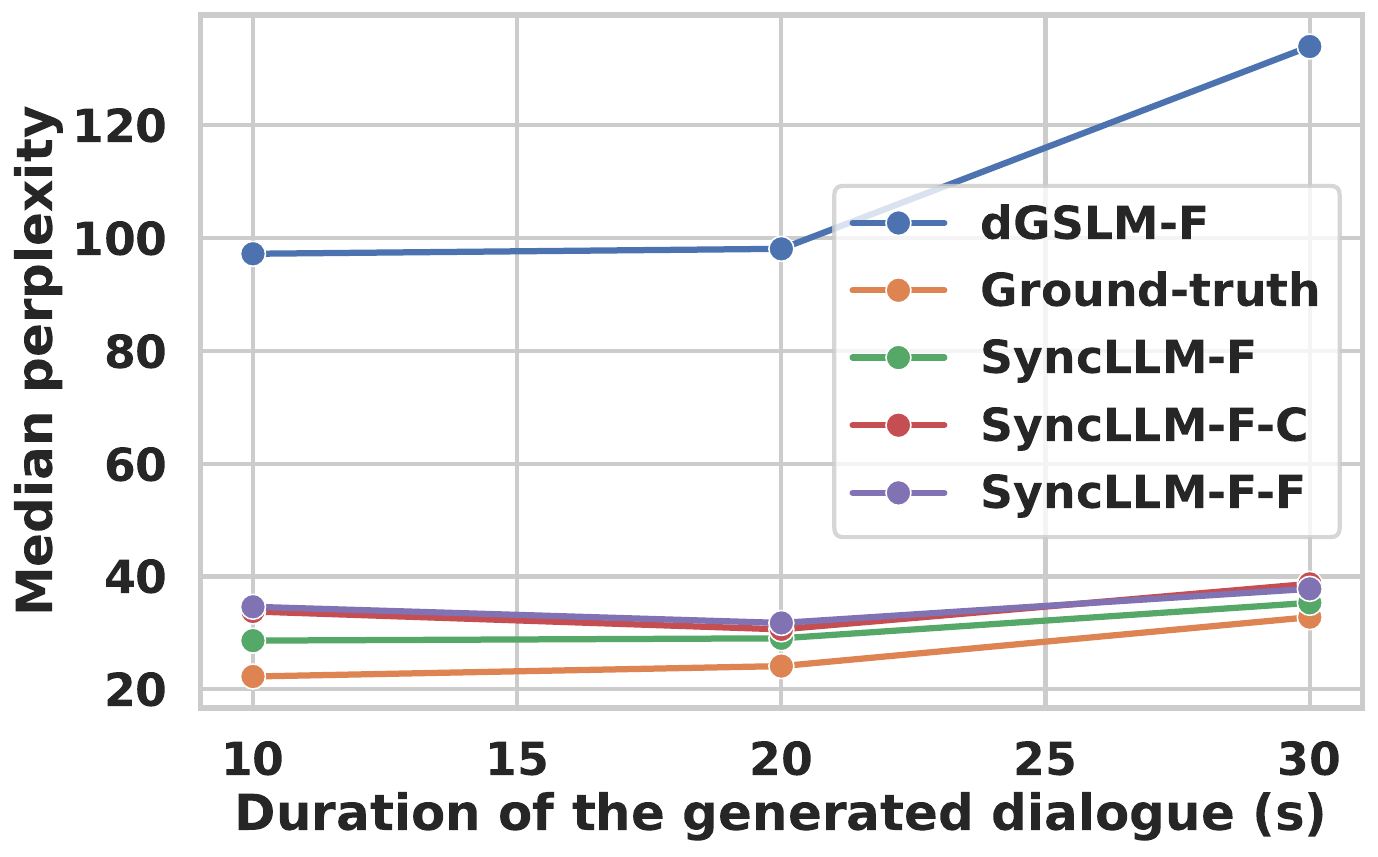}
\vskip -0.1in
\caption{Comparison of ASR perplexity between continuation mode and interaction-mode.}
\label{fig:interaction_comp}
\vskip -0.15in
\end{figure}

\subsection{Full-duplex interaction}
\label{sec:full_duplex_interaction}


We simulate LLM-user interaction using LLM-LLM interaction with one-chunk latency. We evaluate our model trained with different chunk sizes, thus simulating different  latencies. We also train a version of SyncLLM with Candor  training split in the third training stage, and simulate its interaction with the original model trained with only Fisher.

In Fig.~\ref{fig:interaction_comp}, we compare median perplexities obtained with prompts sampled from Fisher and Candor test splits. We also show the perplexity of ground-truth and samples generated in the dialog continuation setting for reference. We find that SyncLLM in the LLM-LLM interaction setting is able to closely match the performance of the continuation setting, and perform significantly better than dGSLM in continuation setting. Furthermore, we find that interaction between instances of SyncLLM trained with Fisher and Candor datasets, respectively is are almost the same signifying that SyncLLM can perform a coherent conversation even when user's side of the conversation is generated by a model trained with a different dataset.

\noindent\textbf{Human evaluation.} 
Table.~\ref{tab:human_eval_2} shows  ratings for dGSLM, the Fisher-trained continuation model, and  LLM-LLM interactions. Results corroborate findings  in \xref{sec:full_duplex_interaction} -- LLM-LLM interactions outperform  dGSLM on M-MOS, but are slightly worse  compared to the single model continuation setting.

\begin{table}[htbp!]
\vskip -0.07in
\caption{Human evaluation results for Meaningfulness (Meaning.) and Naturalness (Nat.) mean estimates and standard errors (in parentheses) across all data.}
\vskip -0.1in
\centering
\begin{tabular}{lcc}
\toprule
\textbf{Model} & Meaning. $\uparrow$ & Nat. $\uparrow$\\
\midrule
dGSLM & 1.55 (0.06) & 3.95 (0.08) \\
SyncLLM-F & 3.74 (0.06) & 3.90 (0.06) \\
SyncLLM-F-C & 3.39 (0.06) & 3.78 (0.06) \\
SyncLLM-F-F & 3.47 (0.06) & 3.72 (0.06) \\
\bottomrule
\end{tabular}
\label{tab:human_eval_2}
\vskip -0.15in
\end{table}

\section{Conclusion}

We present Synchronous LLMs, a novel post-training framework that converts {an auto-regressive} LLM into a full-duplex spoken dialogue agent. Synchronous LLMs outperform state-of-the-art  in dialogue meaningfulness while maintaining  turn-taking naturalness. Finally, by simulating full-duplex  dialogue between two agents, we show robustness to  delayed input from Internet-scale latencies, where the agents do not  have immediate access to the 
speech generated by their users.


\section{Limitations and Risks}

\textbf{Limitations.} The performance of Synchronous LLMs could be further improved in terms of speech quality. Currently, we use a simple HiFi-GAN vocoder for speech synthesis, and higher-quality speech could be synthesized from semantic units with a more advanced speech generator. Moreover, we have not studied expressivity and non-verbal sounds in dialogue such as laughter, which could make the spoken dialogue more human-like. Another limitation is the context length; synchronous LLMs are initialized from Llama-3, and therefore have the same sequence length limit which constrained the long-context modeling in dialogue as well as the use of more expressive multi-codebook tokenizers like EnCodec \cite{defossez2022highfi} that have higher token rate.

\noindent\textbf{Ethical considerations.} The proposed model is intended for spoken dialogue agents. In case of failure, the system might generate inappropriate responses and toxicity mitigation may be needed for speech outputs. As for unintended use, one example is that bad actors misuse the model for online scams. Speech watermarking is one  potential approach to counter abuse of the technology.

\section*{Acknowledgments}

The University of Washington researchers are partly supported by the Meta AI Mentorship program, Moore Inventor Fellow award \#10617, UW CoMotion fund, and the NSF.

\bibliography{references}

\appendix





\section{Additional training details}

\subsection{Hyperparameters}
{We trained SyncLLM with the Llama3-8b's original sequence length 8192. In the first stage, we train with a per-gpu batch size of 1 on 128 A100 GPUs, equivalent to a total batch of 8192 x 128 = 1M tokens. We use a learning rate of $3 \times {10}^{-5}$, with 500 step warmup and train for 40k iterations. In the second stage, we reduce the batch size to 512k tokens, learning rate to $2.2 \times {10}^{-5}$ and warmup steps to 200, and train for 6000 iterations. In the last stage, we train with a batch size of 256k tokens, with a learning rate of $1.5 \times {10}^{-5}$ and 100 warmup steps, for 2000 iterations.}  

\begin{table*}[t]
\centering\footnotesize
\caption{\textbf{Ablation evaluations over interleaving level.} WUGGY, BLIMP, Topic-StoryCloze, and StoryCloze assess the knowledge and capacity of the model in lexical, syntactical, and semantic levels respectively. We report the accuracy based on negative-log-likelihood -- normalized by the number of tokens -- minimization prediction. The tasks are evaluated in the zero-shot setting.}
\begin{tabular}{l c c c c}
\toprule
\textbf{Interleaving} & \textbf{WUGGY}$\uparrow$ & \textbf{BLIMP}$\uparrow$ & \textbf{Topic-StoryCloze}$\uparrow$ & \textbf{StoryCloze}$\uparrow$\\
\midrule
Turn-level &  63.0  & 56.0 & 76.5 & 55.1\\
Sentence-level &  70.3  & 56.3 &  83.0 & 61.8\\
\bottomrule
\end{tabular}
\label{tab:zs_eval}
\end{table*}

\subsection{Benchmarking interleaving strategies}
We explore two text-speech interleaving strategies in stage 1 of our training: i) Sentence-level interleaving: each sentence is chosen randomly to be either text modality or speech modality. ii) Turn-level interleaving: each turn is chosen randomly to be either text modality or speech modality, resulting in consistent modality for all the sentences within the turn. We compare them by evaluating on a set of spoken language understanding benchmarks proposed in \cite{nguyen2020zero}. We report these results in Table \ref{tab:zs_eval}. On these tasks, we observe that sentence-level interleaving outperforms turn-level interleaving across all benchmarks.


\begin{table}[!t]
\centering
\caption{Comparison of average Pearson correlation of turn-taking event durations between generation and ground-truth continuation with SyncLLM in the two-model interaction setting. Measured on testsets comprising both Fisher and Candor testsets.}
\vskip -0.1in
\label{tab:interaction_naturalness}
\begin{tabular}{cccc}
\hline
\textbf{Latency} & \textbf{SyncLLM-F-F} & \textbf{SyncLLM-F-C} \\
\hline
160 ms & 0.32 & 0.36 \\
200 ms & 0.31 & 0.35 \\
240 ms & 0.28 & 0.32 \\
\hline
\end{tabular}
\vskip -0.15in
\end{table}

\begin{table*}[!t]
\centering
\caption{Comparison of Pearson correlation of turn-taking event durations between prompt and generation.}
\label{tab:prompt_corr}
\begin{tabular}{l*{4}{c}c*{4}{c}}
\hline
\textbf{Model} & \multicolumn{4}{c}{\textbf{Fisher (in-distribution)}} & \phantom{|} & \multicolumn{4}{c}{\textbf{Candor (out-of-distribution)}} \\
\cline{2-5} \cline{7-10}
& \textbf{ipu} & \textbf{pause} & \textbf{fto} & \textbf{Average} && \textbf{ipu} & \textbf{pause} & \textbf{fto} & \textbf{Average} \\
\hline
dGSLM & 0.60 & 0.34 & 0.23 & 0.39 && 0.43 & 0.20 & 0.09 & 0.24 \\
SyncLLM-F (160 ms) & 0.69 & 0.34 & 0.35 & 0.46 && 0.64 & 0.12 & 0.24 & 0.33 \\
SyncLLM-F (200 ms) & 0.57 & 0.49 & 0.29 & 0.45 && 0.61 & 0.34 & 0.13 & 0.36 \\
SyncLLM-F (240 ms) & 0.63 & 0.49 & 0.33 & 0.48 && 0.59 & 0.23 & 0.19 & 0.34 \\
\hline
GT & 0.72 & 0.53 & 0.31 & 0.52 && 0.54 & 0.30 & 0.12 & 0.32 \\
\hline
\end{tabular}
\end{table*}

\section{Naturalness-MOS Instructions}
\label{sec:mos_apdx}
Naturalistic turn-taking between two people is characterized by smooth transitions where each participant listens to the other, responds appropriately, and allows for pauses or silences, creating a balanced and dynamic interaction. Typically, the participants try to avoid overlapping speech, although this may occur especially when one participant provides information that they  understood the other by using words like “yeah” or “uh-huh.” Hesitations, pausing, silence, and repairs are also natural events that occur in a conversation between two people.

Here, you will listen to a dialogue between two people and provide a rating for how natural the turn-taking sounds regardless of its content (the meaning of the words used) and the clarity of voices. 

Some of the samples are generated by an AI model, some are actual recordings of humans in conversation, and some are actual recordings of people, but with AI generated voices overlayed.  Please try to assess the naturalness of the turn-taking without taking into consideration the sound of the voices.

To begin, first listen to the “prompt” audio in its entirety. This is the first part of the conversation. Then listen to the “continuation” audio in its entirety. This is the second part of the conversation. Note that in many cases the voices in the prompt may differ from the voices in the continuation (including the perceived gender of the speakers). Your rating should reflect how natural the “continuation” audio sounds given the turn-taking characteristics you observe in the  “prompt.”

\subsection{N-MOS \& M-MOS}
\label{sec:n_mos}
We provide the complete protocol used for human evaluation of turn-taking \textit{Naturalness} and dialogue content \textit{Meaningfulness}.

\textit{Audios presented}

Please base your rating on the impression you have that two people are talking and listening naturally with one-another in the “continuation” audio.
\begin{etaremune}[topsep=0pt,itemsep=-1ex,partopsep=1ex,parsep=1ex]
  \item Excellent - basically indistinguishable from human-like turn-taking
  \item Good - minor differences from human-like turn-taking
  \item Fair - substantial differences from human-like turn-taking
  \item Poor - very little in common with human-like turn-taking
  \item Bad - essentially nothing in common with human-like turn-taking
\end{etaremune}

\subsubsection{Meaningfulness-MOS}
In this task you will listen to a dialogue between two people and provide a rating for how meaningful their conversation is. By meaningful we mean the degree to which the content of the conversation is coherent and plausible (can you understand the intent of the speakers and does it sound like something people would reasonably talk about). Just as in everyday conversations, the content may or may not be perfectly grammatical, but must be understandable in the context of the conversation.

To begin, first listen to the “prompt” audio in its entirety. This is the first part of the conversation. Then listen to the “continuation” audio in its entirety. This is the second part of the conversation. Note that in many cases the voices in the prompt may differ from the voices in the continuation (including the perceived gender of the speakers). Your rating should reflect how meaningful the “continuation” audio is, given the “prompt.”

\textit{Audios presented}

Please base your rating on the impression you have that the continuation is a meaningful “continuation” of the prompt audio - that it represents a plausible direction the conversation would go and is coherent.

\begin{etaremune}[topsep=0pt,itemsep=-1ex,partopsep=1ex,parsep=1ex]
  \item Excellent - all of the conversation content is plausible and coherent
  \item Good - most of the conversation content is plausible and coherent
  \item Fair - some of the conversation content is plausible and coherent
  \item Poor - little of the conversation content is plausible and coherent
  \item Bad - basically none of the conversation content is plausible and coherent
\end{etaremune}

\begin{figure}[t]
\centering
\includegraphics[width=0.85\linewidth]{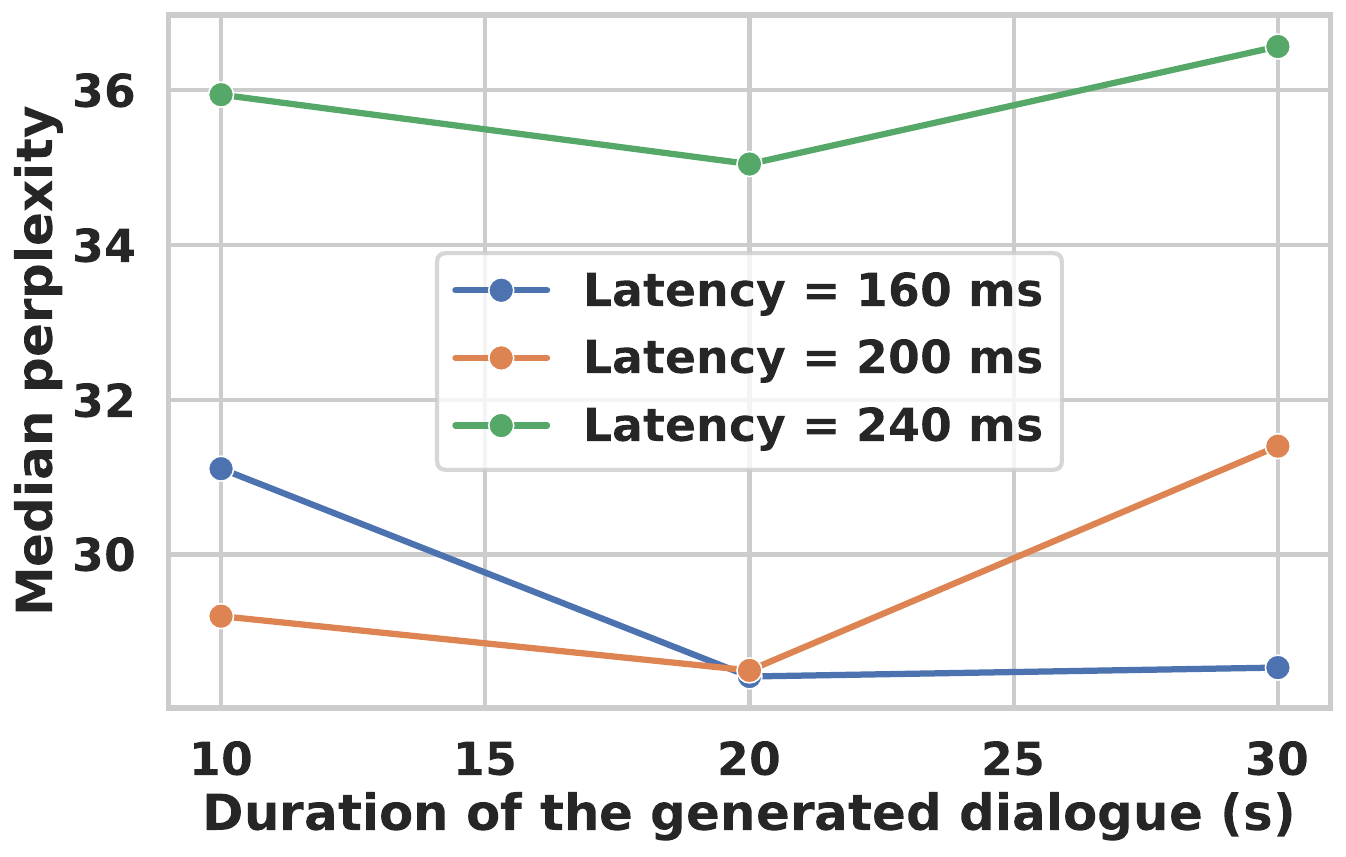}
\vskip -0.1in
\caption{Effect of latency  on two-model interaction.}
\label{fig:interaction_latency}
\vskip -0.1in
\end{figure}


\section{Effect of latency on full-duplex interaction}

In Fig. \ref{fig:interaction_latency}, we compare the performance in the interaction setting with different latencies. We find that our method is robust to a latency as much as 200 ms, but the performance drops with latency greater than that. Similar to our naturalness evaluation in the continuation setting in \xref{sec:naturalness_eval}, to evaluate turn-taking capability of SyncLLM in interaction setting, we compare Pearson correlation of the duration of turn-taking events in generation and ground-truth continuations. In Table \ref{tab:interaction_naturalness}, we observe that on a combined test set of in-distribution and out-of-distribution prompts, performance in the interaction setting closely matches with latencies 160 ms and 200 ms, but drops with 240 ms.

\section{Turn-taking event correlation between prompt and generation}
Similar to the naturalness evaluation in Table~\ref{tab:gt_corr}, where we consider ground-truth continuation as the reference for turn-taking event statistics, we could also consider prompt as the reference. In a way, this measures style consistency between prompt and the continuation. In Table \ref{tab:prompt_corr}, we compare turn-taking event correlation of generations of our method in continuation setting, with that of dGSLM method. We observed that our method demonstrates better turn-taking correlation with the prompts as well.

\end{document}